\documentclass[10pt]{article}

\usepackage{fancyhdr} 

\usepackage[utf8]{inputenc}

\usepackage[utf8]{inputenc}
\usepackage[numbers]{natbib}
\usepackage{comment}

\usepackage{enumitem}
\usepackage{draftwatermark}
\SetWatermarkText{}
\SetWatermarkScale{3}
\SetWatermarkColor[gray]{0.85}
\SetWatermarkAngle{30}

\usepackage{amsmath}
\usepackage{geometry} 
\usepackage{titlesec} 
\numberwithin{equation}{section}
\usepackage[hyphens]{url}  
\usepackage[colorlinks=true, linkcolor=blue, citecolor=blue, urlcolor=blue, breaklinks]{hyperref} 
\usepackage{booktabs} 
\newcommand{\si}[1]{\text{#1}} 

\newcommand{\yr}{yr}                
\newcommand{\yrgamma}{\gamma \ln(2)\,\times\,1\,\si{\yr}}

\newcommand{\wyear}{W\kern 0.1em$\cdot$\kern 0.1em yr} 
\newcommand{\dyear}{\kern 0.1em\kern 0.1em yr\(^{-1}\)} 
\newcommand{\tflopy}{TFLOP\kern 0.1em/\kern 0.1em yr}      
\newcommand{\tflopsw}{TFLOP\kern 0.1em/\kern 0.1em s/\kern 0.1em W}      
\newcommand{\ntoken}{nats\kern 0.1em/\kern 0.1em token} 

\newcommand{\PFLOP}{PFLOP} 
\newcommand{\PFLOPY}{PFLOP\kern 0.1em/\kern 0.1em yr} 
\newcommand{\PFLOPYW}{PFLOP\kern 0.1em/\kern 0.1em yr/\kern 0.1em MW} 
\newcommand{\PFLOPSW}{PFLOP\kern 0.1em/\kern 0.1em s\kern 0.1em/\kern 0.1em MW} 

\newcommand{\MW}{MW} 
\newcommand{\MWYEAR}{MW\kern 0.1em$\cdot$\kern 0.1em yr} 

\usepackage{graphicx} 
\usepackage[font=small, labelfont=bf]{caption} 
\usepackage{array} 
\usepackage{float}
\usepackage{afterpage} 

\usepackage[colorlinks=true, linkcolor=blue, citecolor=blue, urlcolor=blue]{hyperref}

\titlespacing*{\section}{0pt}{10pt plus 2pt minus 2pt}{5pt plus 2pt minus 2pt}

\date{}
\begin{document}

\title{The Race to Efficiency: A New Perspective on AI Scaling Laws}

\author{Chien-Ping Lu\\
\href{mailto:cplu@nimbyss.com}{cplu@nimbyss.com}}

\begin{center}
  \vspace*{0cm} 
  \rule{\textwidth}{2pt} \\ 
  \vspace{0.5cm} 
  {\huge \textbf{The Race to Efficiency: A New Perspective on AI Scaling Laws}} \\ 
   \vspace{0.5cm} 
  {\large \textbf{Chien-Ping Lu}} \\ 
  \vspace{0.1cm} 
  {\large \url{cplu@nimbyss.com}} \\ 
  \vspace{0.2cm} 
  \rule{\textwidth}{0.5pt} \\ 
  \vspace{0.5cm} 
\end{center}

\begin{abstract}
\noindent
As large-scale AI models expand, training becomes costlier and sustaining progress grows harder. 
Classical scaling laws (e.g., Kaplan et al.~\cite{kaplan2020scaling}, Hoffmann et al.\cite{hoffmann2022chinchilla}) predict training loss from a static compute budget yet neglect time and efficiency, prompting the question: 
how can we balance ballooning GPU fleets with rapidly improving hardware and algorithms?
We introduce the \textbf{relative-loss equation}, a time- and efficiency-aware framework that extends classical AI scaling laws. 
Our model shows that, without ongoing efficiency gains, advanced performance could demand millennia of training or unrealistically large GPU fleets. 
However, \textbf{near-exponential progress} remains achievable if the “efficiency-doubling rate” parallels \textbf{Moore’s Law}. 
By formalizing this \textbf{race to efficiency}, we offer a quantitative roadmap for balancing front-loaded GPU investments with incremental improvements across the AI stack. 
Empirical trends suggest that sustained efficiency gains can push AI scaling well into the coming decade, 
providing a new perspective on the \emph{diminishing returns} inherent in classical scaling.
\end{abstract}

\section{Introduction}
\label{sec:intro}

The future trajectory of AI scaling is widely debated: some claim that ever-growing models and datasets are nearing practical and theoretical limits~\cite{sutskever2024reasoning, insider2024orion, time2024progress}, while others maintain that ongoing innovations will continue driving exponential growth~\cite{huang2024scaling, nadella2024scaling, schmidt2024scaling}. For organizations weighing these divergent views, a central question arises: should they “front-load” GPU capacity—relying on the predictable (yet potentially plateauing) gains promised by static scaling laws—or invest in R\&D for (possibly unpredictable and hard-to-measure) efficiency breakthroughs, model innovations, and future hardware enhancements? Ultimately, if diminishing returns do indeed loom, \emph{how severe} might they be in terms of both time and hardware capacity (\emph{see} Table~\ref{tab:five_scenarios_revised} for an illustrative range of outcomes)?

To address this conceptual gap, we note that any truly enduring “exponential” trend hinges on improving an \emph{efficiency} metric that reflects both the outcomes and the costs (time, energy, etc.). Historically, Moore’s Law embodied such progress by showing that transistor \emph{count} per unit area could approximately double every two years~\cite{moore1965cramming}, while Dennard Scaling~\cite{dennard1974design} kept power usage in check. Turning to AI, classical scaling laws quantify how training loss predictably decreases with increasing compute, provided balanced model, data, and training configurations—referred to as the \emph{compute-optimal condition}~\cite{kaplan2020scaling, hoffmann2022chinchilla}. However, these laws are inherently \emph{static}: they do not account for the \emph{severity} of diminishing returns or specify \emph{how quickly} efficiency must improve to offset these trends over time.

\paragraph{Key Idea: Making Scaling Time- and Efficiency-Aware.}
Classical scaling laws~\cite{kaplan2020scaling, hoffmann2022chinchilla} posit that
\(L_0 \propto C_0^{-\kappa}\) for a given \emph{static} compute budget \(C_0\). We extend
this snapshot into a \emph{time- and efficiency-aware} framework. Let \(L_0\) represent the “baseline” loss
associated with an initial compute budget. 
If \(\gamma\) denotes the annual \emph{efficiency-doubling} rate (in \(\text{yr}^{-1}\)), reminiscent of the 0.5 times per year doubling of transistor density in Moore's Law, we derive a \textbf{relative-loss equation} that captures how loss evolves over time:
\begin{equation}
  L(t)\;=\; L_0 \, R(t),
  \quad
  R(t)\;=\;\left(1 + \frac{2^{\gamma t} - 1}{\yrgamma}\right)^{-\kappa}.
  \label{eq:intro_relative_training_loss}
\end{equation}

\begin{figure}[htbp]
  \centering
  \includegraphics[width=\textwidth]{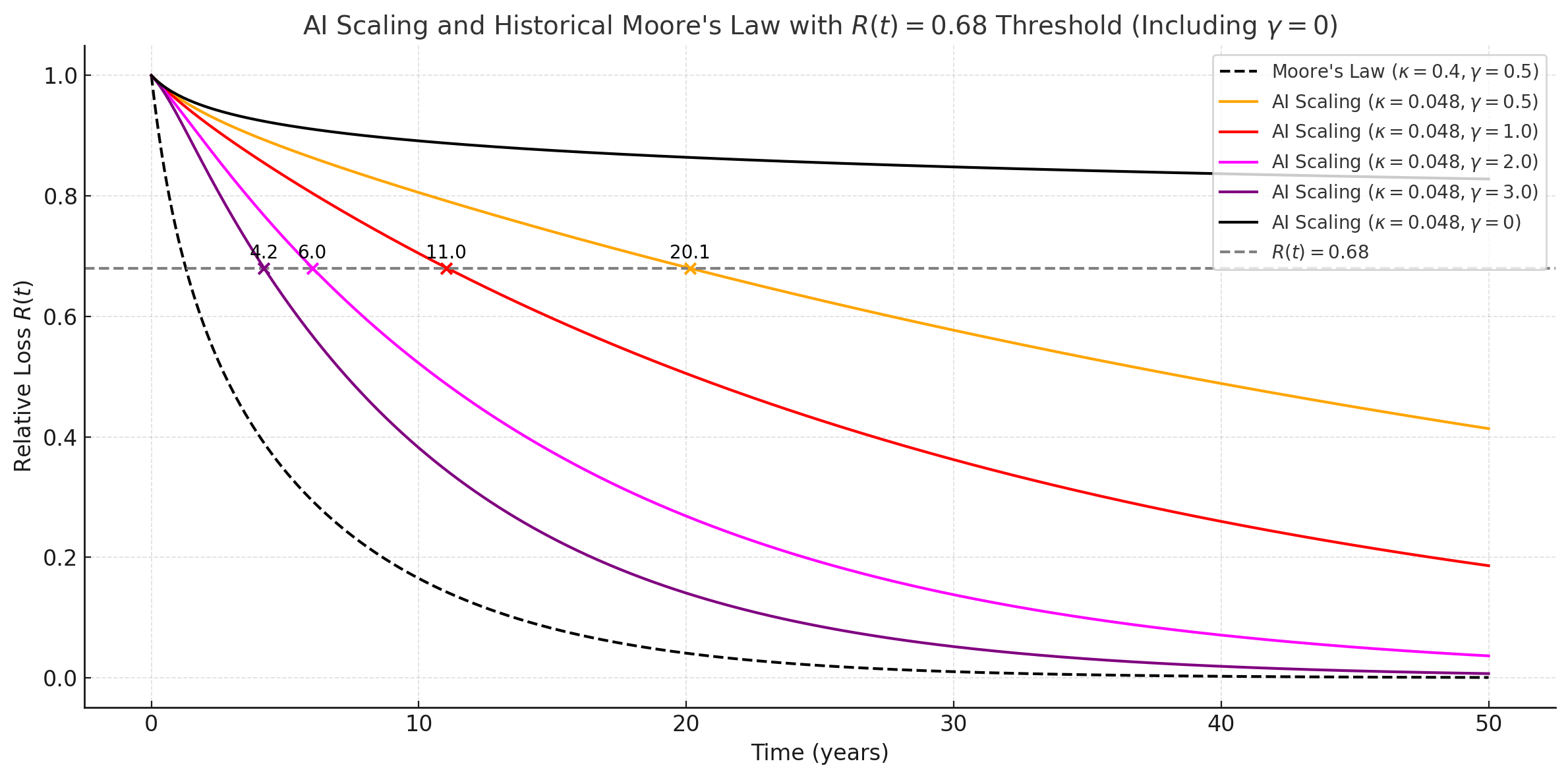}
  \caption{
    \textbf{AI Scaling and Moore's Law with Efficiency-Doubling Rates.}
    This plot compares a hypothetical Moore's Law curve (dashed) with
    \(\kappa = 0.4\) and \(\gamma=0.5\), against AI scaling curves (solid) at
    \(\kappa=0.048\) (typical of large language models) for various
    efficiency-doubling rates \(\gamma\in\{0,0.5,1,2,3\}\). The horizontal line
    \(R(t)=0.68\) corresponds to a token-prediction probability of 50\%, assuming
    \(L_0=1.0\). Increasing \(\gamma\) drastically reduces the time to cross this
    threshold. The x-axis represents \textbf{Time (years)}, and the y-axis represents
    \textbf{Relative Loss \(R(t)\)}. Distinct colors are used for different \(\gamma\)
    values to highlight the impact of efficiency improvements.
  }
  \label{fig:ai_scaling_moores_law}
\end{figure}

Here, \(L(t)\) represents the training loss at time \(t\) (in years), \(L_0\) is the initial loss, \(R(t)\) is the relative loss, and \(\kappa\) is the unitless scaling exponent. The equation captures how training loss evolves over time as efficiency improves. Even with diminishing returns (\(\kappa \ll 1\)), rapid efficiency gains (\(\gamma > 0\)) can sustain near-exponential progress in AI scaling.

\noindent
\textbf{Figure~\ref{fig:ai_scaling_moores_law}} illustrates the interplay between
\(\kappa\) and \(\gamma\). A small \(\kappa = 0.048\) (typical of large language models) causes the AI scaling
curves to flatten significantly over time. With \(\gamma = 0.5\) (efficiency doubling every two years), reducing
\(R(t)\) to 0.68 might require approximately 20 years. Increasing \(\gamma\) to 2.0 compresses
this timeline to well under a decade, while \(\gamma = 3\) shrinks it further. This demonstrates how
higher efficiency-doubling rates can effectively mitigate the limitations imposed by a small \(\kappa\).

By contrast, the \(\gamma = 0\) (flat) curve underscores the severity of diminishing returns, indicating that reaching \(R(t) = 0.68\) could demand 3{,}000\(\times\) the current GPU capacity or training time—scenarios far beyond real-world feasibility (\emph{see} Table~\ref{tab:five_scenarios_revised}).

\paragraph{Organization.}
Section~\ref{sec:related} reviews related work on AI scaling. 
Section~\ref{sec:modeling} formalizes the time-based extension to scaling laws 
and derives the \emph{relative-loss equation}.  
In Section~\ref{sec:analysis}, we examine the scaling behavior predicted by the relative-loss equation. Section~\ref{sec:implications} presents a 
case study comparing “front-loading” GPUs with sustained efficiency improvements and 
discusses broader implications. Finally, Sections~\ref{sec:conclusion}
and~\ref{sec:future} summarize key findings and propose directions for empirical 
validation and future research.

\section{Related Work}
\label{sec:related}

The study of AI scaling laws has become a cornerstone in understanding how training loss decreases as compute increases under optimized configurations. Kaplan et al.~\cite{kaplan2020scaling} introduced the concept of compute-optimal scaling, demonstrating predictable relationships among model size, dataset size, and compute. Brown et al.~\cite{brown2020language} reinforced these findings through the scaling behavior of Large Language Models (LLMs) such as GPT-3. Hoffmann et al.~\cite{hoffmann2022chinchilla} refined the framework in the Chinchilla setting, underscoring the importance of balancing model size and dataset size to achieve compute-optimality. Collectively, these foundational studies provide empirical measurements of scaling exponents and form the basis for much of the work in this domain.

Building on these foundations, recent research has explored additional factors influencing scaling laws. Sardana et al.~\cite{sardana2024beyond} incorporated inference-time compute costs, proposing methods in which smaller models—trained with much larger (potentially synthetic) datasets—can balance efficiency across both training and deployment phases. Snell et al.~\cite{snell2023testtime} investigated strategies for optimizing compute specifically at test time. To address various optimizations, Clark et al.~\cite{clark2022unified} introduced sparsity-aware scaling laws for Mixture-of-Experts (MoE) architectures, formalizing an “effective model size.” Building on that framework, Kumar et al.~\cite{kumar2024scalinglawsprecision} examined precision-aware scaling, showing how precision influences effective parameter counts in a compute-optimal regime.

Despite these advancements, most studies treat compute as a \emph{static} input rather than a \emph{dynamic, time-evolving} resource. This paper addresses that gap by integrating empirically established scaling exponents with the \emph{temporal} dynamics of efficiency improvements, inspired by Moore’s Law~\cite{moore1965cramming} and Dennard Scaling~\cite{dennard1974design}. Our work bridges the gap between classical scaling laws and the real-world constraints of time and efficiency, providing a framework for understanding how diminishing returns can be offset by continuous innovation.

\section{Mathematical Foundation}
\label{sec:modeling}

We now formalize how to extend classical, static AI scaling laws into a dynamic,
\emph{time-dependent} framework. In particular, we derive a \textbf{relative-loss
equation} that unifies traditional loss--compute relationships with a
“Moore’s Law-like” perspective on efficiency gains.

\subsection{Key Parameters and Notation}

Table~\ref{tab:key_parameters} summarizes the main parameters and variables.
In brief, we measure:
\begin{itemize}[label=-, leftmargin=1.5em, itemsep=0.5ex]
    \item \textbf{Logical (Model) FLOPs}, defined to remain stable, vendor-agnostic, 
    and consistent with industry standards of comparing "teraflops" among different 
    precisions such as FP16, BF16, FP8, and FP4;
    
    \item \textbf{Power and time}, representing real-world usage averaged over a suitable
    timescale (\emph{mean-field assumption});
    
    \item A \textbf{scaling exponent} \(\kappa\), which captures how loss decreases
    with total compute, and an \textbf{efficiency-doubling rate} \(\gamma\), 
    quantifying how rapidly “usable compute” can grow per unit time and power.
\end{itemize}
\begin{table}[htbp]
\centering
\begin{tabular}{@{}lll@{}}
\toprule
\textbf{Symbol} & \textbf{Definition} & \textbf{Units}\\
\midrule
\(t\) & Elapsed time since start of training & \si{\yr} \\
\(E(t)\) & Compute efficiency at time \(t\) & \si{\PFLOPYW}\\
\(E_0\) & Baseline/initial efficiency & \si{\PFLOPYW}\\
\(\gamma\) & Annual efficiency-doubling rate & \si{\dyear}\\
\(P(t)\) & Time-varying power usage & \si{\MW}\\
\(P_0\) & Mean-field power usage & \si{\MW}\\
\(C(t)\) & Cumulative compute up to time \(t\) & \si{\PFLOP}\\
\(C_0\) & Initial cumulative compute (snapshot) & \si{\PFLOP}\\
\(\kappa\) & Scaling exponent & (unitless)\\
\(L_0\) & Baseline (initial) training loss & \si{\ntoken}\\
\(L(t)\) & Training loss at time \(t\) & \si{\ntoken}\\
\(R(t)\) & Relative training loss: \(L(t)/L_0\) & (unitless)\\
\bottomrule
\end{tabular}
\caption{Key parameters and variables. Here, “FLOPs” refer specifically
to \emph{logical, or model FLOPs}—i.e., logical operations determined by the model
architecture and dataset. We often measure loss in \texttt{nats/token}, where
\(0.68\,\text{nats/token}\approx 50\%\) prediction accuracy.}
\label{tab:key_parameters}
\end{table}

\subsection{Continuous Efficiency Gains (\texorpdfstring{$E(t)$}{E(t)})}
\label{sec:continuous_eff}

We model \emph{efficiency} as a continuously evolving resource, reminiscent of
how Moore’s Law once described periodic doubling in transistor density. Concretely,
let
\begin{equation}
E(t) \;=\; E_{0}\,\times\,2^{\gamma t},
\quad
\text{(units: \(\si{\PFLOPYW}\))},
\label{eq:E_of_t}
\end{equation}
where \(\gamma\) denotes the annual rate at which efficiency \emph{doubles}, and
\(E_{0}\) is the baseline efficiency at \(t=0\). Although real improvements may
come in discrete jumps, this continuous approximation is mathematically convenient
and mirrors how large-scale phenomena (e.g., population growth) are often modeled
as exponentials.

\paragraph{Relation to Power and Hardware.}
Efficiency, as used here, is dimensionally \(\frac{\text{Logical FLOPs}}{\text{time}\times \text{power}}\).
In practice, raising \(E(t)\) can come from:
\begin{itemize}[label=-,leftmargin=1.5em, itemsep=0.5ex]
    \item \emph{Better hardware} (e.g., next-gen accelerators, lower-precision
    logic, advanced memory or networking),
    \item \emph{Algorithmic gains} (e.g., quantization, expert routing),
    \item \emph{Software optimizations} (kernel-level efficiency, distributed
    training overheads), or
    \item Any combination of the above.
\end{itemize}
We simply aggregate all these factors into a single, time-varying $E(t)$.

\subsection{Cumulative Compute as an Integral (\texorpdfstring{$C(t)$}{C(t)})}
\label{subsec:cumulative_compute}

Classical scaling laws treat compute $C$ as a static budget. Here, we let $C(t)$ \emph{accumulate} over time:
\[
C(t) \;=\; C_0 \;+\; \Delta C(t),
\]
where $C_0$ is the initial snapshot of compute (reflecting prior investments), and
\[
\Delta C(t)\;=\;\int_{0}^{t} E(\tau)\,P(\tau)\,d\tau.
\]
If $P(\tau)$ denotes the power allocated to training, then $E(\tau)\,P(\tau)$ is the instantaneous compute throughput (PFLOP/yr). Integrating from $0$ to $t$ yields the total additional compute \(\Delta C(t)\) beyond the original $C_0$.

\paragraph{Mean-Field Assumption.}
Rather than modeling \(P(\tau)\) at every instant, we approximate
\[
  P(\tau) \;\approx\; P_0,
\]
the \emph{average power} over one year. This “mean-field” approach is common in physics (e.g., average particle collisions) and engineering (e.g., duty cycles). Importantly, this assumption represents an \textbf{upper bound} on performance, as any deviations—such as fluctuations in power usage or suboptimal resource allocation—will result in slower progress in reducing training loss. This makes the mean-field assumption not only mathematically convenient but also practically significant, as it provides an optimistic baseline for evaluating the impact of efficiency improvements. 

For example, consider a training run for LLaMA~3 with 405B parameters, which used approximately \(\,30.8\,\text{million GPU-hours}\) across \(16{,}000\)~H100 GPUs. The \emph{peak} power might reach \(\,16\,\text{MW}\) over a few months. However, spreading this total energy over an entire year yields an \emph{average} power \(P_0\) of approximately \(\,3.5\,\text{MW}\). Instead of modeling short-lived peaks, we “smooth” usage across 12~months to adopt a single constant \(P_0\), simplifying the analysis.

Because classical scaling laws directly link \(C_0\) to \(L_0\), we define
\begin{equation}
  C_0 \;=\; E_0 \cdot P_0 \times 1\,\si{\yr},
  \quad
  \Longrightarrow
  \quad
  \frac{C_0}{E_0\,P_0}\;=\;1\,\si{\yr}.
  \label{eq:C0_invariance}
\end{equation}
Here, \(C_0\) is simply the compute obtained by running efficiency~\(E_0\) at power~\(P_0\) for one year. Changing hardware details (e.g., front-loading more GPUs) merely rescales \(\,(C_0,\,E_0,\,P_0)\), shifting the \emph{initial} loss~\(L_0\) but preserving the \emph{relative} shape of \(L(t)\). Consequently, the time-extended scaling law’s trajectory remains the same, regardless of the precise cluster schedule or deployment plan.

\paragraph{Practical Significance.}
The mean-field assumption represents an \textbf{upper bound} on performance, as any deviations—such as fluctuations in power usage or suboptimal resource allocation—will result in slower progress in reducing training loss. This makes the assumption not only mathematically convenient but also practically significant, as it provides an optimistic baseline for evaluating the impact of efficiency improvements. 

Additionally, the mean-field assumption consolidates \textbf{human R\&D cycles} and \textbf{training sessions}. Even if a specific training run takes only a few months, the development process—including model design, data preparation, and hardware procurement—often spans a year or more. By adopting a one-year baseline, our framework naturally aligns with these real-world timelines, providing a practical and intuitive timescale for planning and evaluation. This consolidation ensures that the relative-loss equation remains relevant across multiple iterations of model development and deployment.

\subsection{Deriving the Relative-Loss Equation}
\label{subsec:deriving_rle}

In the static regime, scaling laws state that the training loss \(L\) decreases
as a power-law of compute, \(L \propto C^{-\kappa}\). Introducing \emph{time}
into the compute accumulation \(C(t)\) transforms this into a time-varying
equation:

\[
  L(t) \;=\; L_0 \,\Bigl(1 + \tfrac{\Delta C(t)}{C_0}\Bigr)^{-\kappa}.
\]

Using the integral form for \(\Delta C(t)\) and noting
\(C_0 = E_0\,P_0 \times 1\,\si{\yr}\), plus the integral
\(\int_0^t 2^{\gamma \tau}\, d\tau = \tfrac{2^{\gamma t} - 1}{\gamma\,\ln2},\)
we obtain:
\[
  \Delta C(t)
  \;=\;
  \frac{E_0\,P_0}{\gamma\,\ln(2)}\,\bigl(2^{\gamma t} - 1\bigr).
\]
Hence,
\begin{align}
  L(t)
  &=\;
  L_0
  \Bigl(1 + \frac{\Delta C(t)}{C_0}\Bigr)^{-\kappa}
  \;=\;
  L_0
  \left(1 + \frac{2^{\gamma t}-1}{\yrgamma}\right)^{-\kappa},
  \label{eq:L_of_t_final}
\end{align}
which can be rewritten in \emph{relative-loss} form:

\begin{equation}
  R(t) \;=\; \frac{L(t)}{L_0} 
  \;=\;
  \left(
    1 + \frac{2^{\gamma t} - 1}{\yrgamma}
  \right)^{-\kappa}.
  \label{eq:relative_loss_equation}
\end{equation}

The \textbf{relative-loss equation} captures how a baseline loss \(L_0\)
evolves over time, provided efficiency improves at a rate \(\gamma\). As
\(\gamma\) increases, \(R(t)\) \emph{declines} more rapidly.

\paragraph{Interpretation.}
\begin{itemize}[label=-, leftmargin=1.5em, itemsep=0.5ex]
  \item \emph{Static vs.\ Dynamic.}
  The \textbf{relative-loss equation} extends \emph{static} scaling laws
into a \emph{time-} and \emph{efficiency-aware} domain. When efficiency does not improve (\(\gamma=0\)), the system effectively
  reverts to “static” scaling. One could, in principle, keep training on the
  same hardware for a very long time, making \(\Delta C\) grow linearly with
  time. 
  \item \emph{Moore’s Law-Like Perspective.}
  By letting “efficiency” double over time (instead of having a single
  snapshot), the analysis aligns with the historical notion of transistor-density
  doubling. Here, \(\gamma\) denotes how quickly one can “refresh” hardware
  and/or optimize software.
\end{itemize}

\subsection{Timescale and Cross-Project Scope}
\label{subsec:timescale_scope}

\paragraph{One-Year Baseline.}
Our derivation adopts a one-year baseline (via the mean-field
assumption~\eqref{eq:C0_invariance}), so $\Delta C(t)/C_0$ measures how
compute accumulates \emph{beyond} that one-year mark. In principle, any
timescale—weeks or months—could be used, yielding the same curve shape; but
one year naturally aligns with budgeting cycles and hardware-refresh periods.
Thus, statements like “doubling efficiency every six months” or “it takes
five years to reduce loss below 0.68” gain clear operational meaning for R\&D
planning.

\paragraph{Multi-Year, Cross-Project Context.}
Although the equations might appear to describe a single, multi-year training
run, organizations typically develop AI systems iteratively across multiple
releases—\emph{upcycling} existing models~\cite{he2024upcycling, vavre2024llama},
refining data pipelines, and introducing new hardware. Each iteration
effectively raises efficiency (\(\gamma>0\)), while \emph{training loss} (e.g.,
cross-entropy) offers a monotonic yardstick: newer models must aim for
\emph{lower} loss to surpass predecessors. In this sense, the
\emph{relative-loss equation} becomes a multi-project roadmap: every new wave
of improvements compounds upon earlier ones, rather than relying on a single,
continuous training job.

\section{Analysis of Scaling Behaviors}
\label{sec:analysis}

Having established a time-based framework for AI scaling, we now examine how its
two principal parameters---the scaling exponent \(\kappa\) and the annual
efficiency-doubling rate \(\gamma\)---shape long-term performance.

\subsection{Reduction to Classical Scaling Laws at \(\gamma=0\)}
\label{subsec:gamma_zero_from_equation}

Starting from the \emph{relative-loss equation}:
\[
R(t)
\,=\,
\Bigl(1 \;+\; \frac{2^{\gamma\,t} - 1}{\yrgamma}\Bigr)^{-\kappa},
\quad
\text{where}
\quad
L(t)\;=\;L_0 \,R(t),
\]
we now set \(\gamma = 0\). To handle the limit 
\(\,2^{\gamma\,t} - 1 \;\to\;\gamma\,t\,\ln(2)\) for small \(\gamma\,t\),
we recall the first-order expansion 
\(\,2^x \approx 1 + x\,\ln(2)\) for \(x \to 0\). Thus,
\[
\tfrac{2^{\gamma\,t} - 1}{\yrgamma}
\;\xrightarrow{\gamma \to 0}\;
\tfrac{\gamma\,t\,\ln(2)}{\yrgamma}
\;=\; t.
\]
Hence, at \(\,\gamma=0,\)
\[
L(t)
\,=\;
L_0 \,\Bigl(1 \;+\; \frac{t}{1\,\si{\yr}}\Bigr)^{-\kappa}.
\]

\paragraph{Interpretation.}
When \(\gamma=0\) (no time-based efficiency improvements), this outcome reduces to the
\emph{static-scaling} form \(L \propto C^{-\kappa}\). However, we now see how running
the \emph{same} hardware and software for an additional time~\(t\) merely accumulates
compute in a linear fashion. As the nearly flat \(\gamma=0\) curve in
Figure~\ref{fig:ai_scaling_moores_law} shows, one must either
\emph{(a)}~train for an exceedingly long duration or
\emph{(b)}~invest in a massive up-front cluster at \(t=0\) to further reduce loss.
Thus, the original diminishing returns \((L \propto C^{-\kappa})\) are now
made explicit in both \emph{time} (\(t\)) and \emph{space} (\(L_0\)), underscoring
why progress inevitably stalls without ongoing efficiency gains (\(\gamma>0\)).

\subsection{Asymptotic Behaviors}

Recall that

\[
R(t) \;=\; \left(1 + \frac{2^{\gamma t} - 1}{\yrgamma }\right)^{-\kappa}
\;\;\Bigl(\text{Equation~\ref{eq:relative_loss_equation}}\Bigr),
\]

and hence,

\[
R(t) \;\propto\; 2^{-\kappa\,\gamma\,t}
\quad\text{for large }t.
\]

Since \(\kappa \gamma > 0\), \(R(t)\) declines exponentially as \(t \to \infty\), mirroring
the vanishing returns one encounters when investing ever more compute. This parallels
the leveling-off observed in Figure~\ref{fig:ai_scaling_moores_law}.

For added intuition, consider a hypothetical analogy to historical Moore’s Law: if one
estimates an effective \(\kappa \approx 0.4\), then a doubling rate of \(\gamma=0.5\)
(doubling roughly every two years) might suffice to maintain improvements for a surprisingly
long time. 

By contrast, \emph{modern} AI scaling laws typically have much smaller
\(\kappa \approx 0.05\). Achieving equally robust gains within a decade may therefore
require \(\gamma\ge2\) (efficiency doubling every six months) or faster. As we increase
\(\gamma\), we effectively prolong what could be termed the “productivity cycle”---the
window in which near-exponential improvements remain viable.

\begin{figure}[h!]
    \centering
    \includegraphics[width=.7\textwidth]{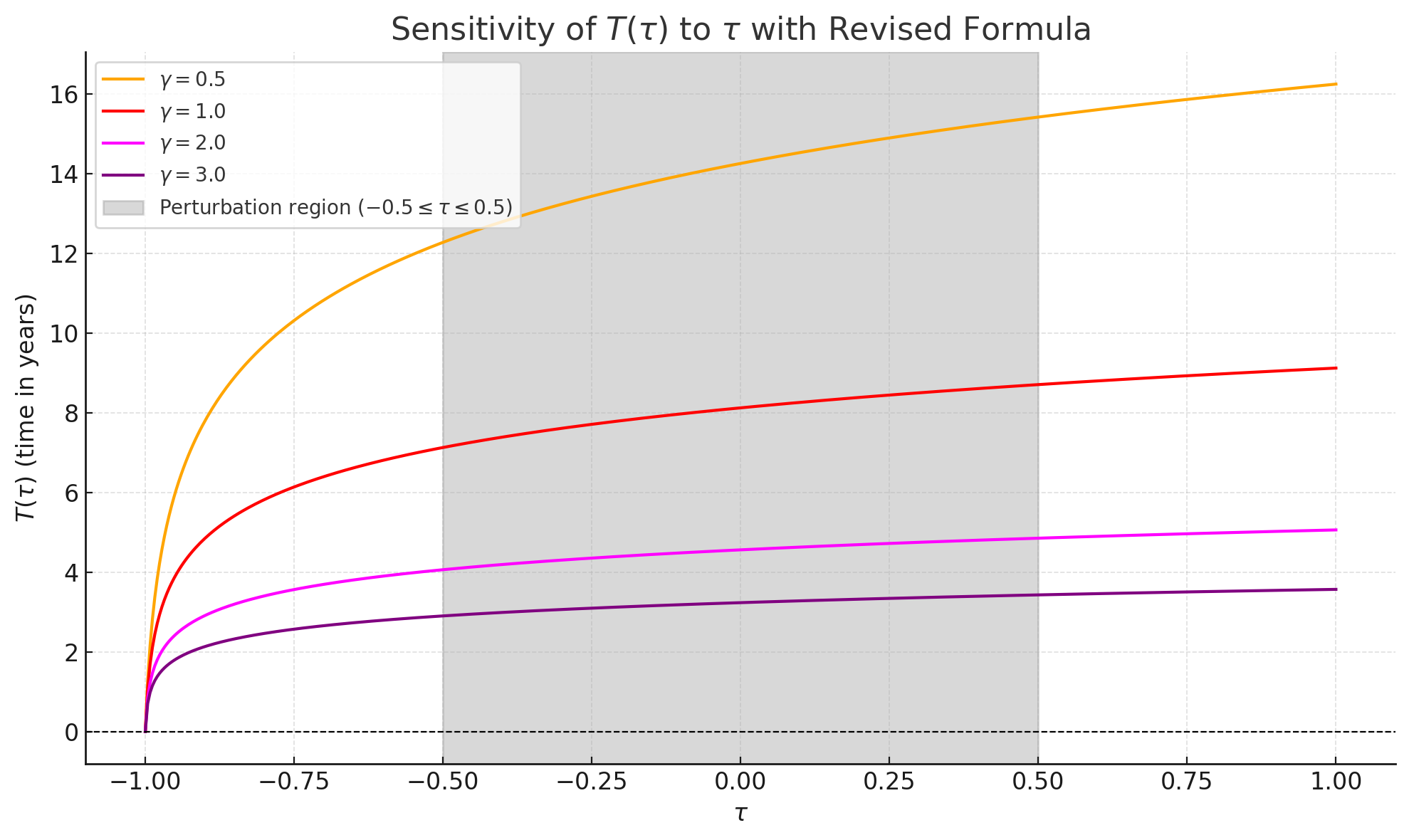}
    \caption{
    \textbf{Sensitivity to baseline perturbations.}
    The horizontal axis shows \(\tau\) in years, with \(\tau=-1\,\si{yr}\) representing
    a scenario where the baseline effectively vanishes. Even under large deviations,
    higher \(\gamma\) values preserve robust predictions for time-to-target.
    }
    \label{fig:tau_sensitivity}
\end{figure}

\subsection{Sensitivity Analysis}

Our model assumes a constant average power budget \(P_0\) under the mean-field assumption
(Equation~\eqref{eq:C0_invariance}). In reality, infrastructure, workloads,
and hardware may fluctuate, introducing uncertainty. To quantify how such changes
affect predictions, we add a perturbation \(\tau\) via

\[
\frac{C_0}{E_0 \cdot P_0} \;=\; 1 + \tau\,\si{yr}.
\]

If \(y\) is a target relative loss (say, \(R(t(\tau)) = y\)), then near \(\tau=0\),
the time-to-target \(t(\tau)\) is approximately \(1/(\gamma\,\ln2)\),
regardless of \(y\). This implies a consistent first-order sensitivity across
different baselines.

For large \(\gamma\), small shifts in effective compute have an even smaller impact on
time-to-target, as illustrated in Figure~\ref{fig:tau_sensitivity}. At, e.g.,
\(\gamma=2\) (\(2\times\) efficiency every six months), the time to reach a moderate
loss threshold changes only slightly when the baseline is perturbed. For example, \(\tau=1\) extends the time-to-target from 5.06 to about
5.78 years—a modest increase for a long-term projection.

\begin{figure}[h!]
    \centering
    \includegraphics[width=.7\textwidth]{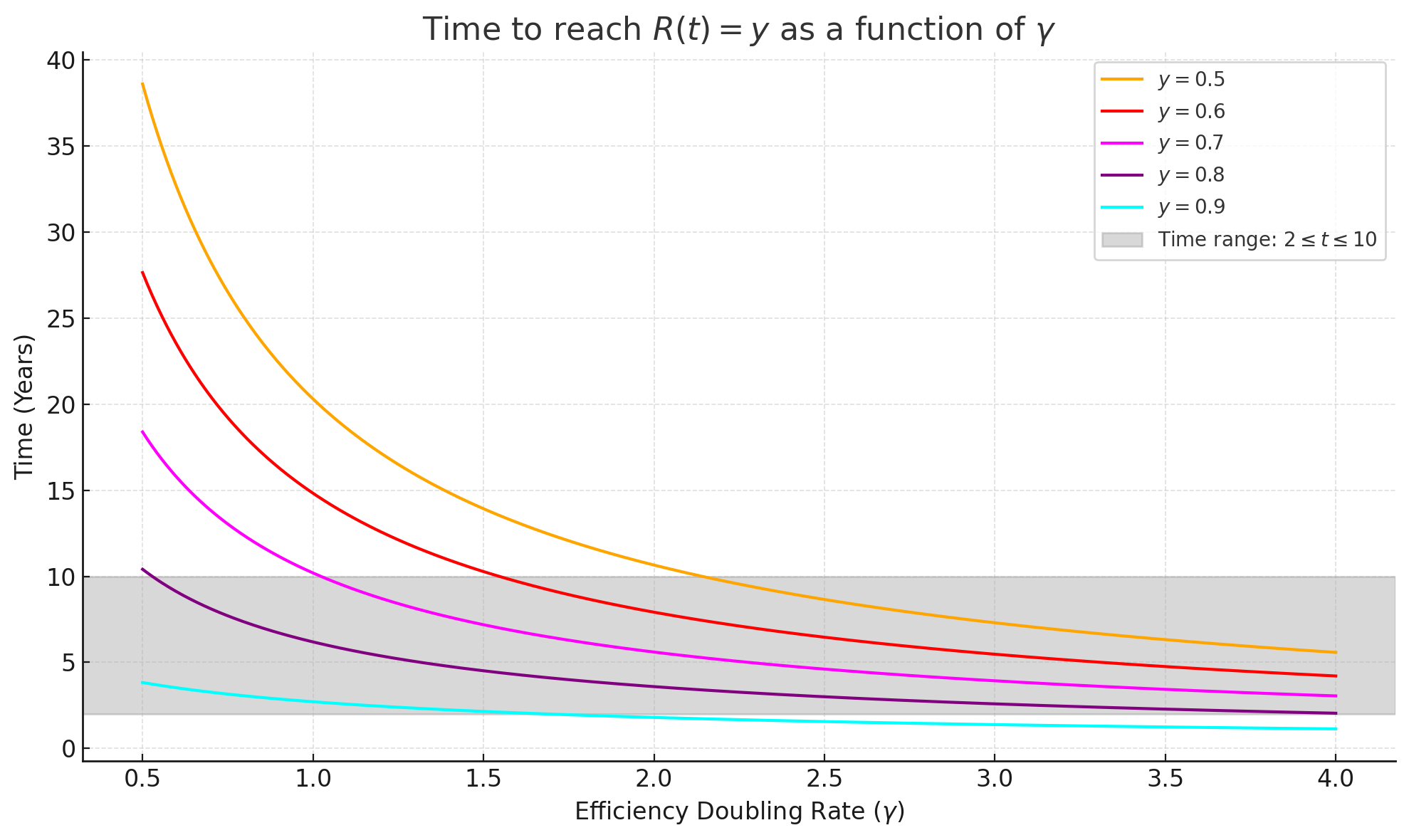}
    \caption{
    \textbf{Time horizons vs.\ efficiency-doubling rate.}
    Higher \(\gamma\) values radically shorten the timelines for achieving targets
    \(y\in [0.5,\,0.9]\). The shaded region (2--10\,yrs) marks a modern industrial
    time frame. Rates \(\gamma\ge2\) align more closely with today’s AI development speeds.
    }
    \label{fig:time_betas}
\end{figure}

\subsection{Efficiency Doubling Rates and Time Horizons}

Finally, consider how changing \(\gamma\) impacts the time needed to attain specific
relative-loss targets. Figure~\ref{fig:time_betas} tracks the time to achieve
various \(y\in [0.5,\, 09]\) for different \(\gamma\) values. From a practical
standpoint:

\begin{itemize}[label=-]
    \item \textbf{Historical Moore’s Law (\(\gamma=0.5\))}:  
    Doubling every two years may suffice for modest goals over long timescales, but
    it can push more stringent targets (e.g., 0.7 or lower) out to 15--20 years—
    far beyond most industry planning cycles.

    \item \textbf{Modern Demands (\(\gamma\ge2\))}:  
    Efficiency doublings every 6--12 months (\(\gamma=2\) or 3) compress
    the entire schedule to a handful of years, in line with contemporary AI’s
    rapid iteration. 
\end{itemize}

These analyses collectively highlight how \(\gamma\) and \(\kappa\) together
determine the feasibility of near-exponential progress. Even small differences
in \(\gamma\) drastically alter the trajectory, underscoring the importance of
continuous innovation.

\section{Implications and Case Studies}
\label{sec:implications}

Having introduced a time- and efficiency-aware perspective on AI scaling, we illustrate its consequences through several \emph{thought experiments}. First, we examine two \emph{theoretically static} scenarios (\(\gamma=0\)), where compute is ``unfolded'' either in \emph{space} (front-loading all GPUs simultaneously) or in \emph{time} (running a fixed-size cluster for millennia). Next, we consider scenarios with positive \(\gamma\), exploring how a balance between up-front GPU investment and sustained efficiency gains can shape multi-year outcomes. Please see Table~\ref{tab:five_scenarios_revised} for the numerical results of these experiments.

\paragraph{Illustrative Scenario.}
To ground these theoretical insights, consider a target of achieving about 50\% token-prediction accuracy (\(L=0.68\,\si{\ntoken}\)). This performance level implies that, on average, the model correctly predicts the next token roughly half the time—no small feat when dealing with large vocabularies and highly nuanced contexts.

Suppose we begin with a model (conceptually \(10\times\)~LLaMA~3 via reorganized feed-forward layers into 16 experts) that trains for one year on \emph{100,000 GPUs}, resulting in an initial loss of \(L_0 = 1.0\,\si{\ntoken}\). Reducing this loss from \(1.0\) to \(0.68\) reflects a substantial improvement in the model’s ability to capture linguistic patterns and make meaningful predictions. Although a literal \(10\times\) increase over a 16,000-GPU cluster would be 160,000 GPUs, we use 100,000 for simplicity and forward-looking assumptions about hardware availability. This scenario illustrates the immense compute resources required to train large-scale models and underscores the trade-offs among model size, training duration, and performance gains.

\subsection{How Severe Are the Diminishing Returns in AI Scaling?}
\label{subsec:static_bignumbers}

A simplified numeric example demonstrates how \emph{purely static} assumptions (\(\gamma=0\)) can result in extremely large hardware requirements or extended training durations, primarily due to the small scaling exponent \(\kappa\) (e.g., \(\kappa = 0.048\)).

\paragraph{From \boldmath$L_0=1.0$ to \boldmath$L=0.68$.}
Under the static law (i.e., the relative-loss equation at \(\gamma=0\)), we have:
\[
\Bigl(1 + \frac{t}{1\,\si{\yr}}\Bigr)^{-\kappa} \;= \; \Bigl(1 + \frac{\Delta C}{C_0}\Bigr)^{-\kappa} \;=\; 0.68,
\]
which implies \(t \approx 3000\) and \(\Delta C \approx 3000 \times C_0\).

\paragraph{Interpretation:}
\begin{itemize}
  \item \textbf{Longer Training (\emph{unfold in time}):}  
  If \(C_0\) denotes running a baseline cluster for one year, reducing loss from \(L_0=1.0\) to \(L=0.68\) under a \(\gamma=0\) assumption would require an additional 3,000 years of training on the \emph{same} hardware.

  \item \textbf{Bigger GPU Fleet (\emph{unfold in space}):}  
  If the loss reduction must be achieved within a single year, \(\Delta C \approx 3000 \times C_0\) implies that the hardware must be scaled by a factor of 3,000. This would push GPU requirements into the \emph{300 million} range, consuming power comparable to the electricity use of an entire continent.
\end{itemize}

\paragraph{Implications: \(\gamma>0\) as Computational Necessity.}
Classical scaling laws, when taken in a purely static sense (\(\gamma = 0\)), imply that driving the loss from \(L_0 = 1.0\) to \(L=0.68\) might require a 3{,}000\(\times\) increase in compute (or equivalently, 3{,}000 years of training on the same hardware). However, this enormous ``3{,}000\(\times\) factor'' does not directly translate to realistic \emph{time} and \emph{resource} constraints.  By introducing \(\gamma>0\) (the efficiency-doubling rate), our framework extends the predictive power of classical scaling laws to more realistic settings, where hardware refreshes, architecture refinements, and data-pipeline optimizations occur continually. Mathematically, one now has
\[
  L(t) \;=\; L_0 \,\Bigl(1 + \frac{2^{\gamma t} - 1}{\yrgamma}\Bigr)^{-\kappa},
  \quad \gamma>0,
\]
so that periodic improvements in ``usable compute'' offset the high cost of further reducing loss. 

In practice, major AI labs do achieve these ongoing gains via hardware refreshes, architecture refinements, and data-pipeline optimizations. Our framework \emph{formalizes} this necessity, preserving the \emph{predictive power} of classical scaling laws while integrating \emph{time} and \emph{efficiency} for more realistic outcomes.
\begin{table}[htbp]
\centering
\begin{tabular}{l|c|c|c|c|c}
\toprule
\textbf{Scenario} 
& \textbf{Initial GPUs} 
& \(\gamma\)
& \(L_0\)
& \(R(t)\)
& \textbf{Time to \(L=0.68\)}\\
\midrule
\textbf{Unfold in Space}  
& \(\sim \mathbf{3\times10^8}\) 
& 0 
& 0.68 
& 1.00 
& \(\sim 1\)\,yr\(^\dagger\) \\[4pt]

\textbf{Unfold in Time}
& 100k
& 0
& 1.00
& 0.68
& \(\sim \mathbf{3{,}000}\)\,yrs \\[4pt]

\textbf{Baseline} 
& 100k 
& 0.5 
& 1.00 
& 0.68 
& \(\sim 20\)\,yrs \\[4pt]

\textbf{Turtle}   
& 10k  
& 3.0 
& 1.12 
& 0.61 
& \(\sim 5\)\,yrs \\[4pt]

\textbf{Hare}     
& 150k 
& 2.0 
& 0.95 
& 0.71 
& \(\sim 5\)\,yrs \\
\bottomrule
\end{tabular}
\caption{
Five illustrative scenarios targeting a training loss of \(L=0.68\) on a model scaled 10\(\times\) beyond the baseline. 
\textit{Unfold in Space} and \textit{Unfold in Time} both set \(\gamma=0\) (strictly static). They differ in whether we pack all needed resources into one year (\(3\times10^8\)~GPUs) vs.\ stretching the same 100k-GPU baseline across millennia. 
By contrast, \textit{Baseline, Turtle,} and \textit{Hare} assume \(\gamma>0\), meaning efficiency improves continuously rather than relying on a single static “snapshot.” 
\\
\(^\dagger\)\,To \emph{reduce} loss \emph{below} 0.68 with \(\gamma=0\) requires exponentially more GPUs, due to the small scaling exponent (diminishing returns).
}
\label{tab:five_scenarios_revised}
\end{table}

\subsection{A Multi-Year Case: Baseline, Turtle, and Hare}
\label{subsec:multi_year_case}

We now consider three illustrative scenarios, each aiming to reduce training loss from an \emph{initial} loss \(L_0\) to a final target of \(L=0.68\). Although they share the same scaling exponent \(\kappa\) and ultimate objective, they differ in:
\begin{itemize}[label=-]
  \item \textbf{Starting Training Loss} (\(L_0\)), which depends on how many GPUs were initially allocated, and
  \item \textbf{Annual Efficiency-Doubling Rate} (\(\gamma\)).
\end{itemize}

In particular:
\begin{itemize}[label=-]
  \item \emph{Baseline}: Begins with \(100{,}000\) GPUs at \(\gamma=0.5\), reflecting a historical rate of doubling roughly every two years.
  \item \emph{Turtle}: Starts with fewer GPUs (\(10{,}000\)) but targets a higher \(\gamma=3.0\) (tripling annually) to see if an “extreme efficiency push” can catch up.
  \item \emph{Hare}: Begins with \(150{,}000\) GPUs (about \(1.5\times\) the Baseline’s cluster) at \(\gamma=2.0\), balancing stronger up-front capacity with a still-robust annual doubling rate.
\end{itemize}

\paragraph{Common Setup and Baseline Loss \boldmath$L_0$.}
We assume each scenario runs for one initial year, after which the training loss is recorded as a new baseline \(L_0\). Formally, if \(L_{\mathrm{init}}\) is the loss before this year of training, and \(\Delta C\) is the additional compute gained in that year (relative to a baseline \(C_0\)), then:
\[
  L_0 
  \;=\;
  L_{\mathrm{init}}
  \,\bigl(1 + \frac{\Delta C}{C_0}\bigr)^{-\kappa}.
\]
Fewer initial GPUs yield a negative \(\Delta C\), as the compute achieved falls short of the baseline \(C_0\), raising \(L_0\). Conversely, more GPUs yield a positive \(\Delta C\), as the compute achieved exceeds the baseline, lowering \(L_0\). This relationship underscores the trade-off between initial resource allocation and the resulting training loss.

\paragraph{Turtle vs.\ Hare in Practice.}

A scenario inspired by DeepSeek-V3~\cite{deepseek2024deepseekv3} could exemplify the \emph{Turtle} approach:
\begin{itemize}[label=-]
  \item \textbf{Smaller GPU Fleet:}  
  Starting with fewer GPUs (e.g., 2,000 rather than 10,000) and aiming for a higher efficiency-doubling rate (\(\gamma>2\)) to offset the lower initial capacity.
  \item \textbf{High-Impact Optimizations:}  
  Even on mid-range devices, leveraging advanced hardware features (e.g., FP8) or specialized software (e.g., Mixture-of-Experts) can systematically increase \(\gamma\).
\end{itemize}
In practice, DeepSeek-V3 has achieved \emph{superior} performance across multiple benchmarks, showcasing how a Turtle-style strategy can deliver both efficiency and state-of-the-art results. Over time, however, growing model or dataset requirements might still necessitate some expansion of the GPU cluster.

By contrast, a Llama~3 (405B)-like~\cite{llama3} scenario may follow the \emph{Hare} approach:
\begin{itemize}[label=-]
  \item \textbf{Front-Loading a Massive Cluster:}  
  Deploying a large fleet (e.g., 30,000 GPUs) to rapidly lower training loss in the first year, reducing initial turnaround but at a high capital cost.
  \item \textbf{Selective Hardware and Architectures:}  
  Potentially a more conservative stance on certain capabilities (e.g., partial adoption of FP8 or limited use of Mixture-of-Experts), with plans to integrate further optimizations in subsequent iterations.
\end{itemize}

In short, these contrasting strategies highlight how organizations may balance a large initial GPU investment (\emph{Hare}) with incremental efficiency gains (\emph{Turtle}) to manage steep diminishing returns. Regardless of fleet size or GPU type, the unifying principle is continuous \emph{efficiency improvement} across hardware, software, and data pipelines—driving the sustained innovation (\(\gamma\)) that underpins modern AI scaling.

\subsection{Scaling Laws as the Driving Force for Innovations}

Sustaining progress in AI scaling demands a deliberate focus on \emph{innovation} to accelerate efficiency gains and counteract diminishing returns. Interestingly, this driving force stems from the same principles that underpin classical scaling laws, once extended into a time- and efficiency-aware framework.

\paragraph{Logical Compute as Optimization-Agnostic.}
Because scaling laws inherently treat \emph{compute} in a model-agnostic manner, \emph{logical compute} is defined as though the model were both dense and full-precision. This prevents conflating \emph{compute} with \emph{efficiency}, the latter reflecting tangible gains from optimizations like sparsity or reduced-precision formats~\cite{kumar2024scalinglawsprecision, clark2022unified}. Without this separation, an architecture such as DeepSeek-V3~\cite{deepseek2024deepseekv3}—which achieves a \(17\times\) higher \emph{real-world} efficiency than Llama~3~(405B) by leveraging sparsity (Mixture-of-Experts), FP8 arithmetic, and other refinements (see Appendix~\ref{sec:compute_optimality})—would appear artificially “smaller.”

By anchoring “compute” to the model’s \emph{full} architectural capacity and attributing actual speedups to \(\mathrm{Time}\times\mathrm{Power}\), the \emph{optimization-agnostic} nature of scaling laws remains intact. Researchers can innovate freely—balancing accuracy, power, and training cost—without altering the fundamental compute measure itself. Meanwhile, the annual efficiency-doubling rate \(\gamma\) quantifies how rapidly these real-world optimizations accumulate, fueling near-exponential progress over multi-year development cycles.

\paragraph{Cumulative Compute as a Compounding Process.}
Classical AI scaling laws (e.g., \(L_0 \propto C_0^{-\kappa}\)) provide a “snapshot” for training loss under a fixed compute budget. Once we generalize to
\[
L(t) \;=\; L_0 \,\Bigl(1 + \frac{\Delta C(t)}{C_0}\Bigr)^{-\kappa},
\]
the additional compute \(\Delta C(t)\) builds over time, transforming that static snapshot into a \emph{compounding} process. Every phase of progress leverages all prior compute investments, allowing near-exponential improvements when efficiency (\(\gamma>0\)) continues rising. This view clarifies why ongoing innovation is indispensable: each incremental gain can magnify the returns of earlier investments, thereby sustaining AI scaling despite inherently small exponents (\(\kappa\)).

\subsection{Outlook}
\label{subsec:takeaways_outlook}

\paragraph{Connection to Industry Trends.}  
Recent industry data highlights significant advancements in AI energy efficiency, with NVIDIA reporting a 45{,}000$\times$ improvement in energy efficiency for AI inference over the past eight years---equating to \emph{doubling approximately every six months} ($\gamma \approx 2$)~\cite{nvidia2024efficiency}. This progress is exemplified by the latest GB10, which integrates the Grace CPU and Blackwell GPU into a desktop form factor~\cite{nvidia2025graceblackwell}. Concurrently, organizations like OpenAI, Google, and Meta continue to refine hardware, software, data pipelines, and infrastructure to sustain rapid improvements in efficiency. For example, DeepSeek-V3~\cite{deepseek2024deepseekv3} achieves substantial efficiency gains through innovations such as FP8 arithmetic, sparsity, and Mixture-of-Experts, demonstrating the potential of architectural optimizations. The success of platforms like Grace Blackwell and innovations such as DeepSeek-V3 underscore the critical importance of prioritizing efficiency improvements in AI development, particularly for organizations with constrained compute resources or sustainability targets. This sustained trajectory of efficiency improvement directly supports the exponential performance gains necessary for training modern large-scale models.

\paragraph{Policy Implications.}
By explicitly embedding the efficiency-doubling rate \(\gamma\) within AI scaling
laws, our framework elevates \emph{innovation} across the AI stack
from an implicit assumption to a measurable driver of progress. Rather than a
“\emph{compute arms race},” AI scaling becomes a \emph{“race to efficiency”}: 
leaders and policymakers can set explicit targets (e.g.\ doubling efficiency
every six months) and synchronize development roadmaps to sustain a high \(\gamma\).
Much as Moore’s Law once provided concrete milestones for transistor scaling,
AI practitioners can now anchor multi-year plans on tangible efficiency
benchmarks, fueling the compounding gains crucial for sustained, real-world AI
advances.

\section{Conclusion}
\label{sec:conclusion}

This work presented a time- and efficiency-aware foundation for AI scaling, 
revealing how a “race to efficiency” naturally emerges once classical, 
\emph{static} scaling laws account for ongoing efficiency gains. Three 
key insights form a mutually reinforcing cycle:

\begin{itemize}[label=-]
    \item \textbf{Time-Extended Perspective.}
    While classical laws link loss to compute at a single snapshot, recognizing 
    “efficiency doubling” over months or years shifts that static view into a 
    dynamic, near-exponential trajectory.

    \item \textbf{Efficiency-Centric Focus.}
    Because scaling laws reliably map compute to performance gains, the 
    crucial question becomes how \emph{efficiently} compute accumulates over 
    time. In this light, \emph{efficiency doubling} emerges as both a 
    computational and practical necessity to mitigate steep diminishing returns.

    \item \textbf{Innovation as Core to Scaling.}
    Once efficiency is central, continual optimizations across the AI stack 
    no longer appear as external “fixes” but as integral parts of the scaling 
    process. These incremental improvements compound over multiple training 
    cycles and product generations, reinforcing the time-extended perspective.
\end{itemize}

Looking ahead, constraints akin to those that once challenged transistor scaling 
may ultimately call for new paradigms. Yet the tension between \emph{diminishing 
returns} and \emph{time-extended efficiency gains} will likely remain a defining 
force in AI’s technological evolution.

\section{Limitations and Future Work}
\label{sec:future}

While the relative-loss equation offers a unified perspective on AI scaling progress, its practical
value and generality warrant further exploration. Below, we highlight several avenues for extending
and refining this framework.

\subsection*{Empirical Validation and Transparency}

The insights in this paper rest on theoretical constructs and empirically observed scaling exponents.
Nevertheless, comprehensive validation against real-world data is crucial. Greater transparency
in reporting relative training loss, cumulative compute, and efficiency-doubling rates could
enable more rigorous cross-study comparisons. Industry-wide data sharing, standardized benchmarks,
and consistent evaluation protocols—akin to those once used for guiding semiconductor progress—would
help verify the predictive power of the relative-loss equation. Such efforts could also inform
resource-allocation decisions, model architectures, and targeted efficiency improvements.

\subsection*{Generalizing to Multi-Phase Growth}

Although our framework focuses on AI training, it could be generalized to capture early,
sub-exponential “kick-off” phases or logistic transitions in other domains—ranging from
technology adoption to broader economic processes. Such a generalization would provide a unified view
of how systems evolve from an initial ramp-up to potentially exponential (or S-curve~\cite{foster1986innovation})
growth trajectories.

\subsection*{Optimal Usage vs.\ Raw Compute}

The relative-loss equation assumes \emph{compute-optimal} usage, where model size, dataset size,
and training strategy are balanced to fully exploit available resources. Simply increasing GPUs or
peak compute does not guarantee improved performance if scaling principles are not followed.

For instance, training a large model without adequate data or failing to tune hyperparameters
may not yield the expected loss reductions. Likewise, imbalanced scaling between model and data can
prevent the envisioned gains. The equation thus reflects a \emph{best-case} trajectory, assuming that
each increment in compute efficiency translates directly into effective training progress.

\subsection*{Extending the Framework to Inference and Test-Time Scaling}

While our current formulation focuses on \emph{training} dynamics, extending the
relative-loss equation or developing analogous constructs for \emph{inference}-time
scaling~\cite{snell2023testtime, wu2024inference} could yield a more holistic view of
AI system performance. In particular, because large models are increasingly used to
\emph{generate} new training data and perform on-the-fly or iterative refinement, higher inference efficiency can \emph{accelerate} the training pipeline rather than merely reduce deployment costs.

As a result, understanding how inference-time efficiency improvements translate into faster
throughput, lower latency, or expanded data pipelines may be crucial, given that
deployment considerations (and the downstream feedback loop into training) increasingly
shape large-model design choices. A unified framework that addresses both training and
inference could thus clarify how hardware roadmaps, data engineering, and algorithmic
optimizations interact across the entire lifecycle of AI systems.

\subsection*{Evolving Concepts of Compute-Optimality}

Traditionally, compute-optimal scaling assumes a \textit{static dataset} and a \textit{fixed model
configuration}. In reality, both datasets and models evolve over time. Approaches such as
\textit{upcycling pretrained models}~\cite{komatsuzaki2022sparse, he2024upcycling, vavre2024llama},
and dynamically adapting \textit{model size or precision} introduce new optimization strategies.
Similarly, dataset generation and curation—leveraging synthetic data~\cite{zelikman2022star} or
reasoning-based selection~\cite{zelikman2024quietstar}—blur the boundary between \textit{model development}
and \textit{data sourcing}.

As these strategies mature, future frameworks must treat \textit{datasets, models, and compute budgets}
as interconnected and evolving. Such adaptability ensures the \textit{relative-loss equation} and
similar scaling models remain relevant.





\appendix

\section{Invariance of Logical Compute}
\label{sec:compute_optimality}

This appendix explains why we define \emph{logical compute} as dense and
full-precision and how it underpins a fair comparison---illustrated with
DeepSeek-V3 and Llama~3 (405B)---even when actual GPU hours differ
across hardware or optimization strategies.

\subsection*{Why \(\kappa\) Remains Unchanged by Optimizations}

Suppose the training loss depends on model size \(N\) and dataset size \(D\) as
\[
L(N, D) \;=\; A\,N^{-\alpha} \;+\; B\,D^{-\beta} \;+\; E,
\]
where \(\alpha,\beta>0\). For a fixed compute budget \(C\), the \emph{compute-optimal}
pairs \((N^*, D^*)\) satisfy
\[
N^* \;\propto\; C^{\tfrac{\beta}{\alpha + \beta}},
\quad
D^* \;\propto\; C^{\tfrac{\alpha}{\alpha + \beta}},
\]
so substituting back yields
\[
L(C)\;\propto\;C^{-\tfrac{\alpha}{\alpha+\beta}}
\;=\;
C^{-\kappa},
\quad
\kappa \;=\; \frac{\alpha}{\alpha + \beta}.
\]
Hence, \(\kappa\) depends only on \(\alpha,\beta,\) not on how we optimize model
parameters or numeric formats. In other words, whether one uses low-precision
arithmetic, sparsity, or a Mixture-of-Experts design, the fundamental exponent
\(\kappa\) stays invariant.

\subsection*{Defining Logical Compute}

We define \emph{logical compute} as though the model is both \emph{dense} and
\emph{full-precision}. Specifically:
\[
\text{Logical Compute (FLOPs)} \;=\; 6 \times N \times D,
\]
where
\begin{itemize}
  \item \(N\) = total model parameters,
  \item \(D\) = total training tokens,
  \item The factor 6 accounts for forward/backward passes and parameter updates.
\end{itemize}
This ensures that “compute” consistently reflects the \emph{full} architecture,
independent of sparsity or precision. Actual speedups (e.g., from FP8 or MoE)
appear \emph{separately} in reduced \(\mathrm{Time}\times\mathrm{Power}\), rather
than shrinking the fundamental compute measure.

\subsection*{Case Study: DeepSeek-V3 vs.\ Llama 3 (405B)}

To illustrate why logical compute is kept dense, consider two models that
achieve broadly similar large-scale \emph{outcomes} yet differ in real-world
GPU usage:

\begin{itemize}
  \item \textbf{DeepSeek-V3}~\cite{deepseek2024deepseekv3}:
    \begin{itemize}
      \item \(N \approx 671\,\mathrm{B}\) (parameters),
      \item \(D \approx 14.8\,\mathrm{T}\) (training tokens),
      \item Real GPU usage: \(\sim 2.78\,\mathrm{M}\) GPU-hours on H800,
        adjusted to \(\sim2.224\,\mathrm{M}\) GPU-hours at H100 equivalence.
    \end{itemize}
  \item \textbf{Llama 3 (405B)}~\cite{llama3}:
    \begin{itemize}
      \item \(N \approx 405\,\mathrm{B}\),
      \item \(D \approx 2.0\,\mathrm{T}\),
      \item Real GPU usage: \(\sim 30.84\,\mathrm{M}\) GPU-hours on H100.
    \end{itemize}
\end{itemize}

\paragraph{Summarizing the Computations.}
\vspace{-0.5em}
\begin{table}[H]
\centering
\resizebox{0.98\textwidth}{!}{%
\begin{tabular}{@{}lcccccc@{}}
\toprule
\textbf{Model} &
\textbf{Parameters} \textbf{(B)} &
\textbf{Data Tokens (T)} &
\textbf{Logical Compute} \textbf{(PFLOPs)} &
\textbf{GPU Hours} \textbf{(M, H100-Eq.)} &
\textbf{Relative Efficiency} \\
\midrule
\textbf{DeepSeek-V3} 
 & 671 & 14.8 
 & \(5.95\times10^{15}\) 
 & 2.224 
 & \(\approx 17.0 \ (\text{vs.\ Llama=1})\) \\
\textbf{Llama 3 (405B)} 
 & 405 & 2.0 
 & \(4.86\times10^{15}\) 
 & 30.84 
 & 1.0 (baseline) \\
\bottomrule
\end{tabular}
} 
\caption{\textbf{DeepSeek-V3 vs.\ Llama 3 (405B).} 
Both achieve large-scale performance but differ in GPU hours. Logical compute
(assuming dense, full-precision) is high in both cases, yet DeepSeek-V3
real-world efficiency is about \(17\times\) Llama 3’s.}
\label{tab:ds_llama_comparison}
\end{table}
\vspace{-1em}

As shown in Table~\ref{tab:ds_llama_comparison}, \emph{logical compute} for
DeepSeek-V3 is slightly larger than Llama 3, reflecting its extra parameters
and bigger training set. However, real-world GPU hours for DeepSeek-V3 are far
\emph{lower} than Llama 3, thanks to hardware/software optimizations (e.g.,
sparsity, FP8). Defining compute in a purely “dense” sense prevents conflating
those optimizations with the model’s \emph{intrinsic} size.

\paragraph{Relative Efficiency.}
One can define a “relative efficiency” factor:
\[
\text{Relative Efficiency} 
\;=\; \frac{\text{Logical Compute (FLOPs)}}{\text{GPU-Hours (H100 Eq.)}},
\]
then normalize Llama 3 (405B) at 1.0. Under that measure, Table~\ref{tab:ds_llama_comparison}
shows DeepSeek-V3 is about \(17\times\) more efficient. Crucially, this ratio
does not shrink its \emph{logical compute} (it remains at
\(\,5.95 \times 10^{15}\,\mathrm{PFLOPs}\)), but records gains via fewer GPU
hours.

\subsection*{Counterarguments and Responses}

\paragraph{Should We Adjust Logical Compute for Sparsity or Precision?}
Some suggest \emph{reducing} FLOPs to reflect only the fraction of parameters
activated per token (\emph{experts-per-token} in MoE) or the lower numeric cost
(e.g., from FP8). However, that merges two concepts:
\begin{itemize}
  \item \textbf{Full Model Complexity:} The entire parameter space at full precision,
        representing the model’s theoretical capacity.
  \item \textbf{Real-World Efficiency Gains:} Achieved by using only a subset
        of parameters, or fewer bits per operation, thus lowering time and power.
\end{itemize}
Conflating them penalizes architectures that are inherently more efficient (like
Mixture-of-Experts). Instead, we \emph{keep} the definition of ``logical compute'' dense
and record any real-world speedups in the denominator (\(\mathrm{Time}\times\mathrm{Power}\)). 
This way, a model reaps the benefits of advanced routing or quantization 
(\(\gamma > 0\)) without artificially reducing its fundamental FLOP count.

\paragraph{Is “Effective Model Size” More Accurate?}
Although some frameworks define an \emph{effective} size \(N_{\mathrm{eff}}\) for
MoE~\cite{clark2022unified} or reduced precision~\cite{kumar2024scalinglawsprecision}, 
such an approach can hide the full parameter space. Not all parameters are 
\emph{active} per token, but they still \emph{exist}, providing capacity for 
generalization and future scaling. By keeping ``logical compute'' dense, we preserve 
fairness across architectures. If a model invests in sophisticated routing or quantization, 
the advantage should appear as a \emph{lower} \(\mathrm{Time}\times\mathrm{Power}\), 
not by discarding parameters from the total.

\subsection*{Summary}

Because \(\kappa\) depends solely on the \emph{power-law slope} linking model
size, dataset size, and compute, it is invariant to whether a model employs
sparsity, lower-precision arithmetic, or mixture-of-experts routing.
Defining \emph{logical compute} as though it were both dense and full-precision
provides a consistent baseline for comparing very different architectures.
Real-world \emph{efficiency} gains, meanwhile, show up in 
\(\mathrm{Time}\times\mathrm{Power}\), thereby highlighting the genuine 
speedups achieved by hardware or software improvements. This framework 
allows us to preserve the foundational exponent~\(\kappa\) from classical 
scaling laws while giving proper credit for engineering advances in 
accelerators, memory systems, or training algorithms.

\end{document}